# Comparative Analysis of Different Methods for Classifying Polychromatic Sketches

**Fahd Baba**
LoLSketch
fahd31256@gmail.com

**Devon Mack**
LoLSketch
devonpmack@gmail.com

**Abstract**

Image classification is a significant challenge in computer vision, particularly in domains humans are not accustomed to. As machine learning and artificial intelligence become more prominent, it is crucial these algorithms develop a sense of sight that is on par with or exceeds human ability. For this reason, we have collected, cleaned, and parsed a large dataset of hand-drawn doodles and compared multiple machine learning solutions to classify these images into 170 distinct categories. The best model we found achieved a Top-1 accuracy of 47.5%, significantly surpassing human performance on the dataset, which stands at 41%.

## 1. Introduction

Handwriting recognition is a mature field in machine learning and computer vision with many open source datasets such as MNIST [1], Quick Draw! [2] and TU-Berlin [3] made publicly available to facilitate research in this area. As a result, many researchers have developed algorithms to analyze and find biases in this data and use this information to either model, categorize, or even generate more data samples. These models prove to be very useful in fields such as robotics where text recognition is vital to the performance of these machines. A large portion of such algorithms use simple neural network architectures to detect and learn patterns across several thousand data points and utilize the learned parameters to make predictions on unseen data. The most common approach being Convolutional Neural Networks (CNN) for image classification and Recurrent Neural Networks (RNN) for vectorized doodle generation [4]. However, these datasets do not provide color or tonality information about the sketches which could give crucial insight about their categorization. In this paper, we build on top of previous research by training and comparing the performance of multiple well established image classification models on a dataset containing colored hand-drawn doodles of characters from the video game League of Legends.

## 2. Related Work

Classifying vectorized image drawings is not a revolutionary idea and presents some challenges notably subjective interpretation and abstraction as observed by Bandyopadhyay et al. in their paper Do Generalised Classifiers really work on Human Drawn Sketches? [5]. People from different backgrounds have different interpretations of words and tend to draw differently depending on variables such as drawing time and drawing ability. As part of their research, they condition the Contrastive Language-Image Pre-training (CLIP) model "by learning sketch specific prompts using a novel auxiliary head of raster to vector sketch conversion". They then use a codebook learning approach to make predictions about the model, this allows the model to interpolate between different levels of abstraction as some sketches may be more detailed than others. Using this method, they obtain a Top-1 accuracy of 77% on the TU-Berlin dataset.

Other papers have developed models specialized for doodle classification such as the work of Guo et al., that presents a new algorithm KNN++, an improvement on the KNN algorithm where multiple centroids were used per class to help identify different features within the same category [6]. This method, though novel and robust, did not yield a high enough accuracy to compete with CNN classifiers. They improved their model by introducing a weighted voting system and tested it on Google's Quick Draw! dataset [2]. The KNN++ Weighted scored a Mean Average Precision (MAP) of 35% while their CNN implementation achieved a score of 60%.

The model that currently holds the state of the art performance in doodle classification is called DeepSketch 3 developed by Seddati et al. [7]. Their novel CNN architecture achieves an accuracy of 79% on the TU-Berlin dataset and 93% on the sketchy database. These results demonstrate impressive performance especially given the variability and abstraction in hand-drawn sketches.

## 3. Data

### 3.1. Collection

Before we begin developing the model, we need a solid dataset with enough training samples to tune the parameters of our algorithm. Datasets such as TU-Berlin and Quick Draw! don't contain color information within the strokes, which is the challenge we are attempting to tackle in this paper. For this reason, we collected League of Legends character doodles from the online drawing game LoLSketch [8]. In about a week, we obtained roughly 112,000 drawings ranging across 170 different characters. We also retrieve the label of each character being drawn, as well as the percentage of players who correctly guessed the drawing. This information will prove to be useful when parsing and cleaning the dataset.

### 3.2. Analysis

As users tend to skip on drawing certain characters, some classes are largely underrepresented with a minimum of 384 samples on Udyr, a maximum of 596 on Cassiopeia, and an overall average of 506 images per class. To further investigate this potential issue, we visualize the data samples by plotting the number of classes for each sample size, and after doing so, we notice the distribution is skewed to the right. This class imbalance could introduce a bias towards overrepresented categories during training, leading to poor generalization particularly in high parameter spaces [9]. In order to make up for this, we randomly resample images from underpopulated classes such that all classes have an equal probability of being selected.

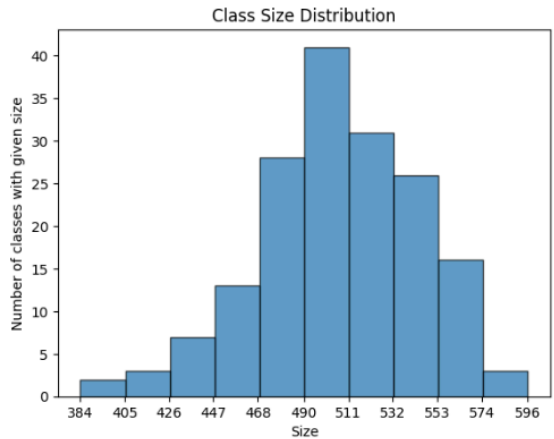

Figure 1: A histogram with 10 bins that plots the frequency of class sizes across multiple ranges. The x-axis represents the number of samples per class, and the y-axis represents the number of classes that fall within each bin.

### 3.3. Cleaning

Because drawings were not moderated during the data collection phase, a significant number of either empty or inappropriate images remain within the dataset. To make the cleaning process easier, we use the number of correct guesses for any given drawing to assess the quality of the sample. Using this metric, we can quickly discard most of the noise in the dataset. After the cleaning, we are left with approximately 80,000 clean training samples.

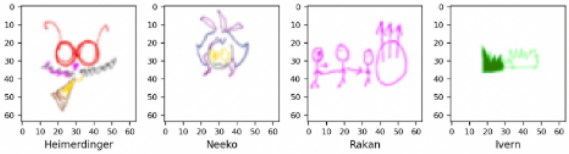

Figure 2: Four colored sketches of league of legends characters with their respective labels displayed underneath each drawing.

## 4. Methodology

### 4.1. CNN

As CNNs such as ResNet [10], MobileNet [11], and VGG [12] are widely considered strong baselines for image classification, we train these popular and well documented architectures to determine which one works best for this task. Since these models have already been trained on a much larger dataset, ImageNet [13], we fine-tune the networks by first downloading and using their pretrained weights as the initial starting point before we begin training. The process of using the weights of a network that has been trained on a larger more complete dataset to further train it on a much more unique and specialized dataset is called fine-tuning and has proven to be very promising and yields better results than training from scratch. This also helps prevent issues such as overfitting and speeds up time to convergence [14].

We also train a small architecture from scratch to use as a baseline for comparison. The network is made up of 4 convolutional blocks, each block is composed of a convolutional layer with a kernel size of 3, stride of 1 and padding of 1. We apply the non-linear ReLU activation after each convolution for computational efficiency and better gradient computation. Following each convolutional layer, a max pooling layer with a kernel size of 2 and stride of 2 is applied, reducing the spatial dimensions of the feature maps by approximately half. This reduction in spatial dimensions enables us to double the number of filters in successive convolutional blocks without significantly increasing the model's size or computational cost. Finally, we flatten the output of the last convolutional layer and add a fully connected layer with one output per class. Since this is our own architecture, pretrained weights are not available for this network, thus we initialize the layers using the popular Kaiming Initialization (also known as He Initialization) [15].

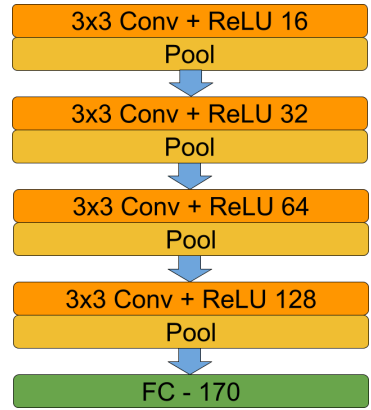

Figure 3: A detailed diagram illustrating the CNN architecture used in this paper.

In order to assess the quality of the trained models, we use metrics aligned with the application of this neural network. In this case, the model will be making live predictions on user drawings until successfully guessing the label. For this reason, we calculate the Top-5 accuracy using the following equation to evaluate the model's performance.

$$\text{Top-N Accuracy} = \frac{1}{N} \sum_{i=1}^{N} \mathbb{I}\left(y_i \in \hat{Y}_i\right)$$

We don't care so much about the order of the predictions than we do about the confidence and the accuracy of the correct label occurring in the top N guesses. These metrics are much easier to interpret in the context of our application as using the accuracy metric, we are able to estimate the average number of guesses required until the model guesses correctly with a given probability.

### 4.2. Clusterability

Each character has distinct features that clearly distinguishes them from the pool of sketches, such as hair color, skin tone, and weapon type. A combination of these attributes is likely enough to differentiate between all 170 classes. However, not all drawings in the dataset follow this strict pattern of highlighted features. Furthermore, we notice several distinct yet common patterns within the same category of drawings. As multiple features alone may give away a character being drawn, and not everyone picks the same feature to draw, this results in many different common drawings within the same class.

For this reason, we theorize there may be multiple clusters of drawings within each category. Similar to the work of Guo et al. in their paper Quick, Draw! Doodle Recognition [6], we attempt to use K-Means++ clustering to find these sub-clusters inside each category and use them to improve the predictions of the K-Nearest Neighbors algorithm. However, we build on their work by first extracting image feature vectors using models that were pretrained on the famous ImageNet dataset and running the clustering algorithm on those features instead of the images themselves. Research has shown that within the feature space of a trained image classification neural network, similar images tend to be closer to each other in euclidean distance [16].

In order to determine the best architecture and weights to use, we calculate the Silhouette Score of the clusters found by K-Means++ on the feature space of each of the models. A Silhouette Score is a metric that is used to measure the quality of clusters in a dataset. In our case, we want to measure how well clustered are the feature vectors produced by our trained models. We can calculate this silhouette score for any given label, in our case, since our data points are already split into multiple classes, we interpret each class as a separate cluster.

$$s_i = \frac{b_i - a_i}{\max(a_i, b_i)}$$

Where $a_i$ is the mean distance from the i-th element to all other points in the same cluster, and $b_i$ is the average distance to all points in the closest cluster. The overall silhouette score can be computed by averaging the silhouette scores of each element.

$$\text{Silhouette Score} = \frac{1}{N} \sum_{i=1}^{N} s_i$$

| Architecture | Silhouette Score |
| --- | --- |
| MobileNetV3Small | −0.05 |
| VGG16 | −0.064 |
| ResNet50 | −0.145 |

Table 1: Each model architecture and its respective Silhouette Score across the dataset using class labels as the true cluster for each example.

The silhouette score ranges from negative one to positive one. Numbers closer to zero generally indicate that the data points tend to be in the boundary between all other clusters, meaning clusters overlap and there's no clear distinction between each one. Negative numbers indicate the data points are closer to incorrect clusters while positive numbers indicate data points are correctly placed within their designated cluster. In order to visualize this, we use t-Distributed Stochastic Neighbor Embedding (t-SNE) [17], a dimensionality reduction algorithm that allows us to project data points from a high dimensional space such as the image feature vectors onto a much lower dimensional space that we can use to visualize the clusters.

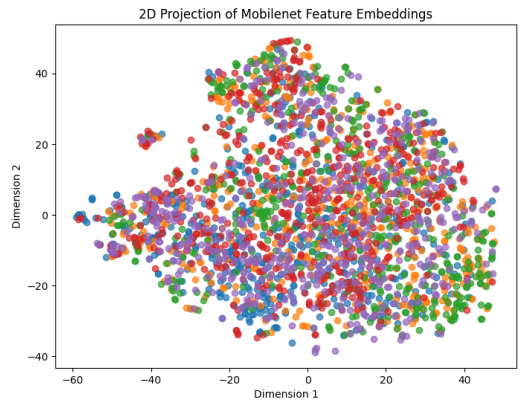

Figure 4: A scatter plot that shows data point clusters of the embedding vectors generated by mobilenet on the lolsketch polychromatic doodle dataset.

### 4.3. KNN++ with Feature Embeddings

Traditional KNN algorithms rely on a distance measurement between the input data and training data to classify images. In order to achieve high accuracy, it is crucial the data points fed into the KNN algorithm are structured in such a way that similar points are closer together forming clusters around each class center. The work of Guo et al., proposes a new algorithm which they name KNN++ that aims to solve the issue of multiple possible drawings of the same class discussed in the previous section [6]. They implemented their algorithm by using the individual pixels as the data points of each image. We theorize using feature embeddings from a pretrained image classification model would yield a better separated feature space with improved clusterability and thus allow for a higher accuracy.

### 4.4. Embedding Space Structure

If the KMeans algorithm properly finds clusters and our feature extractor network is able to understand visual concepts at a high level, we should observe that the clusters group similar images together. To test this hypothesis, we applied KMeans++ on the embedding vectors of images within each class to identify sub-clusters within each category. We then calculated the distance between each embedding vector and the cluster centers and sorted them by euclidean distance to their nearest centroid. We take the 4 closest images to each cluster and plot them vertically. Each column corresponds to one cluster and images are sorted by distance from top to bottom. Because we don't initially know the number of clusters contained in each class, we try different values for k until we find something that splits the data well.

We find that the cluster represents exactly two different types of drawings. It appears some people like to simply draw an abstract version of the character such as the object or thing they represent, in this case, Maokai is a tree, explaining the tree drawings in the dataset. There are still various drawings of the character however, each with a different art style and level of detail, making it difficult to determine the ideal number of clusters.

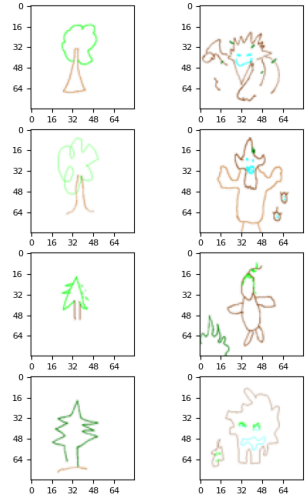

Figure 5: A plot that shows the 4 closest images to 2 different centroids in the Maokai class.

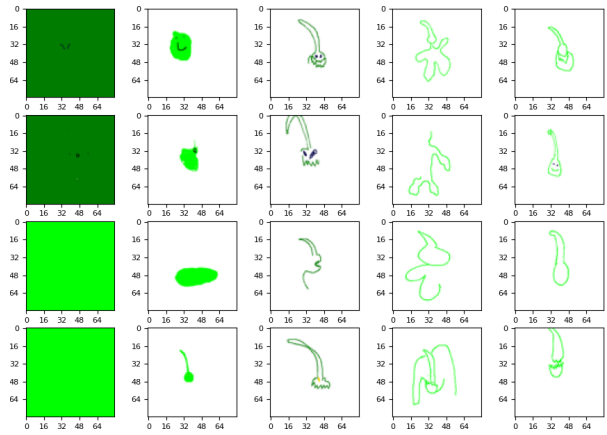

Figure 6: A plot of 4 images closest to 5 centroids from the Zac class plotted vertically where each column represents a different cluster.

Furthermore, it is unlikely all categories have the same underlying number of clusters. By visually inspecting the data, we have established the Maokai class is composed mainly of 2 different clusters. However, there are classes such as Zac that contain significantly more than that. This poses a significant challenge because choosing an incorrect number of clusters k weakens the performance of our image classification algorithm. Using a variable number of clusters per category may introduce unfairness as classes with more centroids will be more prominent in the latent space, increasing their chances of being selected. Although this could be remedied by weighting the influence of each centroid, determining the best weight coefficient remains a non-trivial task. For this reason, we use a fixed number of 3 centroids per class, which we determine by observing the images in each class and intuitively estimating the average number of potential clusters in each class.

## 4.5. Voting Scheme for Image Classification

To classify a given image $x_i$, we use the same voting scheme as in the original implementation. A rank weighted voting system in which given a set of centroids $C$, sorted in ascending order in Euclidian distance to the image $x_i$, the weight $w_i$ for a centroid $c_i$ is given by $\frac{1}{\sqrt{d_i}}$. Where $d_i$ is the distance from the sample $x_i$ [6]. Furthermore, we try different combinations of the number of clusters in the k-means algorithm, and number of neighbors in the KNN algorithm.

## 5. Results

### 5.1. Training Statistics

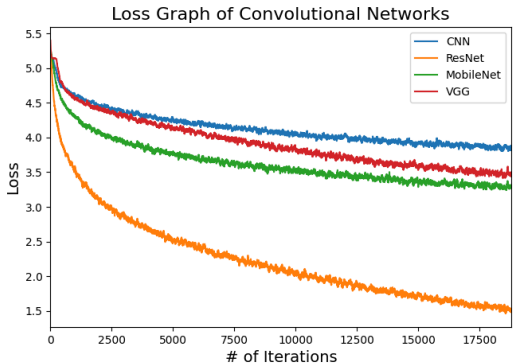

Figure 7: Training loss for different networks smoothed using exponential moving average(EMA).

After training and testing the convolutional neural networks, and experimenting with various hyperparameters, we find that using a CrossEntropy loss with a learning rate of 1 x $10^{-4}$ and a batch size of 16 yields the best results across all architectures. Each model is trained for 35 epochs, with a checkpoint selected based on the lowest validation loss.

### 5.2. Testing and Evaluation Metrics

|  | Top-1 | Top-5 | Top-10 |
|---|---|---|---|
| CNN | 25.3% | 49.8% | 61.3% |
| MobileNet | 33.7% | 56.5% | 66.1% |
| ResNet50 | 47.5% | 69.3% | 76.5% |
| VGG16 | 33.5% | 55.9% | 66.8% |

Table 2: Comparison of the Top-1, Top-5, and Top-10 accuracies of ResNet, MobileNet, VGG, and our custom tiny CNN model.

Since ResNet has the highest accuracy, we use it to compute the confusion matrix. Due to the large number of classes, we display only the five most confused classes in the confusion matrix. However, class sparsity in the test set makes it difficult to compute reliable statistics. To address this, we instead compute the matrix using training data. This is reasonable, as none of the models show signs of overfitting, and their losses and accuracies remain consistent across training, validation, and test sets.

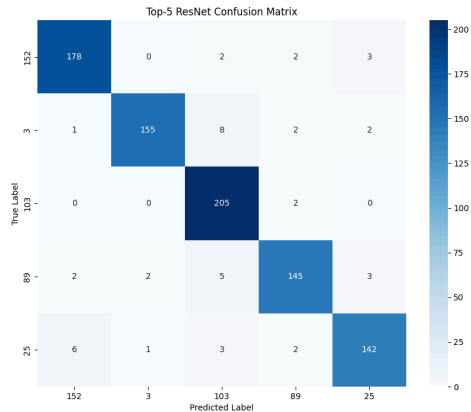

Figure 8: Confusion matrix of the five classes on which ResNet performed the worst.

### 5.3. Clustering Performance with Pre-Trained Networks

The scatter plot in Figure 4 reveals randomly distributed data points with no clear structure, suggesting poor clustering ability. While some dis-tortion may stem from the dimensionality reduction algorithm, the low silhouette score confirms that networks pre-trained ImageNet struggle to cluster our dataset effectively. This is likely due to the high subjectivity and abstraction of the sketches, even images within the same category vary significantly in shape, size, and quality, making it difficult for models to learn high-level concepts from the images.

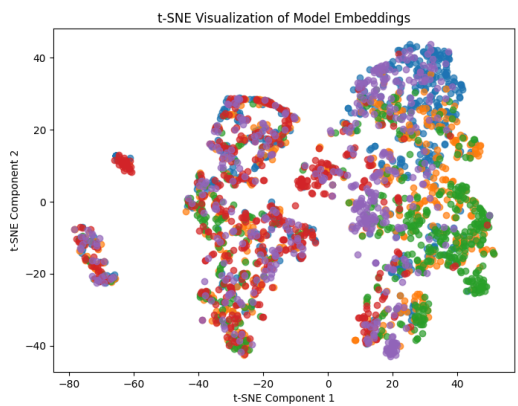

Figure 9: A scatter plot of the image embedding vectors computed with the VGG16 network architecture.

### 5.4. KMeans with Weighted Voting

The feature embeddings produced by the VGG16 network are clearly separated into 4 different clusters. The right cluster seems to have an unusual zig zag shape similar to the left cluster with higher density. Judging by the distribution of the points in the cluster, we notice embeddings are particularly concentrated at different areas, namely the bottom right, top right and all along the left. This however appears to be the only segmented cluster with the others being made up of a fairly even distribution of the labels all through their area.

This poor clustering ability directly impacts model performance, as reflected in accuracy scores. The highest Top-1 accuracy is achieved by the pretrained VGG16, despite its relatively low silhouette score. However, after observing Figure 9 it is clear why VGG16 outperforms the other models. The feature embeddings form distinct clusters, where points from similar classes are positioned closer together. This structured clustering likely explains its superior classification performance.

| Architecture | **Top-1 Accuracy** |
|:---:|:---:|
| ResNet50 | 5.9% |
| MobileNet | 6.8% |
| VGG16 | 9.6% |

Table 3: Top-1 accuracy of KNN++ with feature embeddings extracted from different neural network models.

## 6. Conclusion

In conclusion, we found that the ResNet50 was the neural network that achieved the highest accuracy scores out of all the techniques we have attempted. We also notice it is fairly computationally efficient and not very expensive to train, and achieves a lower loss much faster than its other networks. However, its cluster ability does not outperform the ones of traditional convolutional neural networks such as VGG16, signifying it is likely not the best choice to use with techniques that rely heavily on a clean feature embedding space.

Due to the level of subjectiveness and abstraction in each drawing, and hidden subcategories within each class, KNN struggles to identify appropriate clusters and achieves a relatively low accuracy score. Nevertheless, we speculate this can be greatly improved by choosing a different number of clusters for each class, allowing the centroids to represent much more distinct features in the dataset.

## 7. Acknowledgements

Fahd Baba: Neural network training and testing, KMeans and KNN implementation, image feature extraction

Devon Mack: Data collection, Website integration

## 8. Code

https://github.com/lolsketch/lolsketch-ai-guesser